\documentclass{article}
\usepackage{ijcai21}

\usepackage{times}
\usepackage{soul}
\usepackage{url}
\usepackage{color}
\usepackage[hidelinks]{hyperref}
\usepackage[utf8]{inputenc}
\usepackage[small]{caption}
\usepackage{graphicx}
\usepackage{amsmath}
\usepackage{amsthm}
\usepackage{booktabs}
\usepackage{algorithm}
\usepackage{algorithmic}
\usepackage{graphicx} 
\usepackage{float} 
\usepackage{subfigure} 
\usepackage{stfloats}
\usepackage{bbding}
\usepackage{makecell}
\title{Knowledge-aware Contrastive Molecular Graph Learning}

\author{
Yin Fang$^1$\and
Haihong Yang$^1$\and
Xiang Zhuang$^1$\and
Xin Shao$^2$\and
Xiaohui Fan$^2$\And
Huajun Chen$^1$\\
\affiliations
$^1$College of Computer Science and Technology, Zhejiang University\\
$^2$Pharmaceutical Informatics Institute, College of Pharmaceutical Sciences, Zhejiang University\\

\emails
\{fangyin,zhuangxiang,xin\_shao,fanxh,huajunsir\}@zju.edu.cn,
capriceyhh@gmail.com
}
\begin{document} 

\maketitle
\begin{abstract}
Leveraging domain knowledge including fingerprints and functional groups in molecular representation learning is crucial for chemical property prediction and drug discovery. When modeling the relation between graph structure and molecular properties implicitly, existing works can hardly capture structural or property changes and complex structure, with much smaller atom vocabulary and highly frequent atoms. In this paper, we propose the Contrastive Knowledge-aware GNN (CKGNN) for self-supervised molecular representation learning to fuse domain knowledge into molecular graph representation. We explicitly encode domain knowledge via knowledge-aware molecular encoder under the contrastive learning framework, ensuring that the generated molecular embeddings equipped with chemical domain knowledge to distinguish molecules with similar chemical formula but dissimilar functions. Extensive experiments on 8 public datasets demonstrate the effectiveness of our model with a 6\% absolute improvement on average against strong competitors. Ablation study and further investigation also verify the best of both worlds: incorporation of chemical domain knowledge into self-supervised learning.
\end{abstract}

\section{Introduction}

Molecular representation learning plays an important role in fundamental tasks in the pharmaceutical industry (e.g. drug discovery and molecular property prediction), in replacement of manually designing and extracting fingerprints or physicochemical descriptors as molecular features. In light of deep learning methods, molecular representation learning captures domain knowledge implicitly from massive molecular datasets over years of tedious experiments in a faster and cheaper way.

Since molecules are essentially graphs with atoms and bonds, graph representation learning can be naturally introduced, aggregating and transforming structural information. Recent efforts apply GNNs to map molecular graphs into neural fingerprints, which retains diverse attributes and structural features in the original graphs. GNNs capture sophisticated graph substructures by recursively aggregating its $k$-hop neighboring nodes then compressing into a fixed-size node embedding. 

Despite their strong representation power, one of the limitations of GNNs is \textbf{the lack of domain knowledge}. We observe that pre-trained GNN with large amount of data plus massive GPU training hours is only competitive with k-nearest neighbors algorithm ($k=10$) based on molecular fingerprint similarity on seven molecular property prediction tasks, indicating that chemical domain knowledge such as molecular fingerprint or functional groups is crucial for learning discriminative molecular representation.

Take the halogenation reaction replacing Chlorine by Fluorine $(-Cl \rightarrow -F)$ on the benzene ring as an example, this simple replacement reaction challenges GNNs in three ways, shown in Table \ref{reaction}. Firstly, not all nodes are aware of this small change if GNNs are not deep enough (e.g. 1-hop). In fact, using 1-hop GNN, the Carbon atom in red can not perceive the change before and after the reaction. Secondly, without domain knowledge, GNNs based on neighborhood information can hardly learn meaningful molecular representation for later prediction. Because a small change in the functional group results in a big difference of chemical property between the reactant and the product. Thirdly, our example requires at least 3-hop GNNs to build a latent representation corresponding to the concept of the benzene ring. Note that using deep GNNs may cause the over-smoothing issue \cite{liu2020towards} since GNNs are highly dependent on the ($k$-hop) neighborhoods\cite{feng2020graph}. One way to ease these issues is to fuse domain knowledge into GNNs explicitly.

\begin{table*}[!ht]
\renewcommand\arraystretch{1.5}
  \centering
  \begin{tabular}{p{2.664cm}<{\centering}|p{5cm}|p{5.3cm}|p{3cm}<{\centering}}
    \hline  
   Issue Type & \makecell*[c]{Structure-based GNN}&\makecell*[c]{Contrastive Knowledge-aware GNN}& \makecell*[c]{Illustration}\\
    \hline
      Structural Changes &
        Distant nodes are not aware of atom or functional groups changes. & Contrastive learning from similar molecule pairs based on fingerprint similarity. &
        \begin{minipage}[b]{0.36\columnwidth}
            \centering
            \raisebox{-.9\height}{\includegraphics[width=\linewidth]{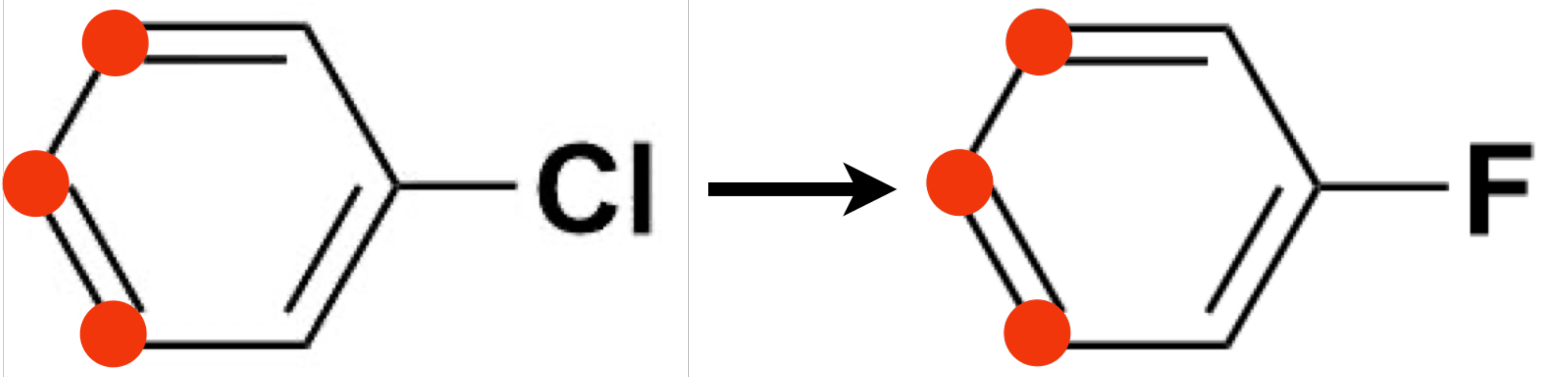}}
        \end{minipage} \\[1pt]
    \hline
        Property Changes &
        Implicitly modeling the relation between structural changes and property changes. &Introducing chemical domain knowledge as a type of inductive bias. &
        \begin{minipage}[b]{0.36\columnwidth}
            \centering
            \raisebox{-.9\height}{\includegraphics[width=\linewidth]{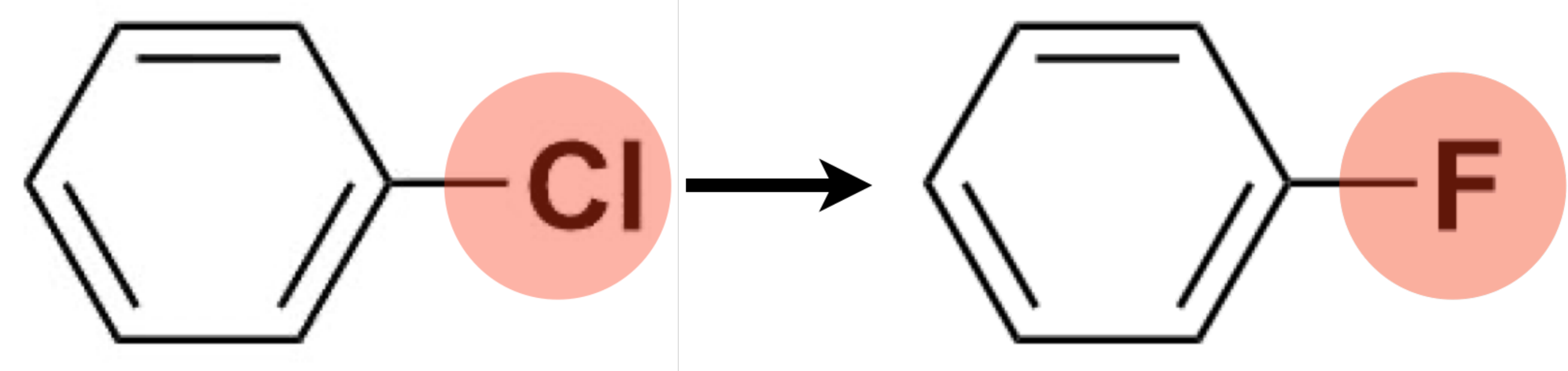}}
        \end{minipage} \\
    \hline
        Complex Structure &
        Deeper GNN is required by complex structure such as benzene ring. & Explicitly fusing functional groups as domain knowledge. &
        \begin{minipage}[b]{0.10\columnwidth}
            \centering
            \raisebox{-.65\height}{\includegraphics[width=\linewidth]{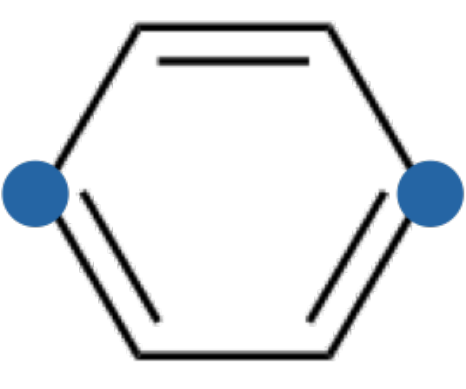}}
        \end{minipage} \\
    \hline
            Highly Frequent Atoms &
        Hard to encode highly frequent atoms appeared in most molecules such as Carbon.& Distinguishing same types of atoms in different functional groups. &
        \begin{minipage}[b]{0.25\columnwidth}
            \centering
            \raisebox{-.95\height}{\includegraphics[width=\linewidth]{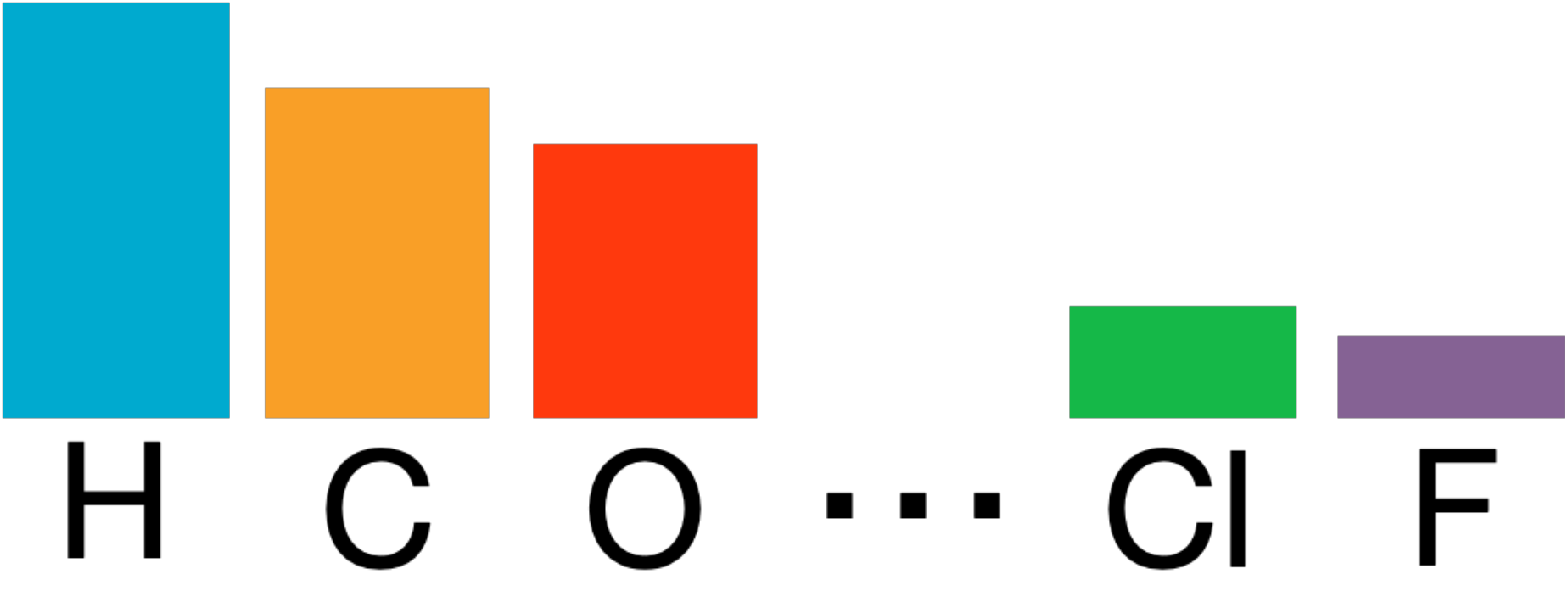}}
        \end{minipage} \\
    \hline
     Small Vocabulary &
        There are only 118 types of atoms in Periodic table. & Relieving representation ambiguity with contrastive learning. &
        \begin{minipage}[b]{0.28\columnwidth}
            \centering
            \raisebox{-.75\height}{\includegraphics[width=\linewidth]{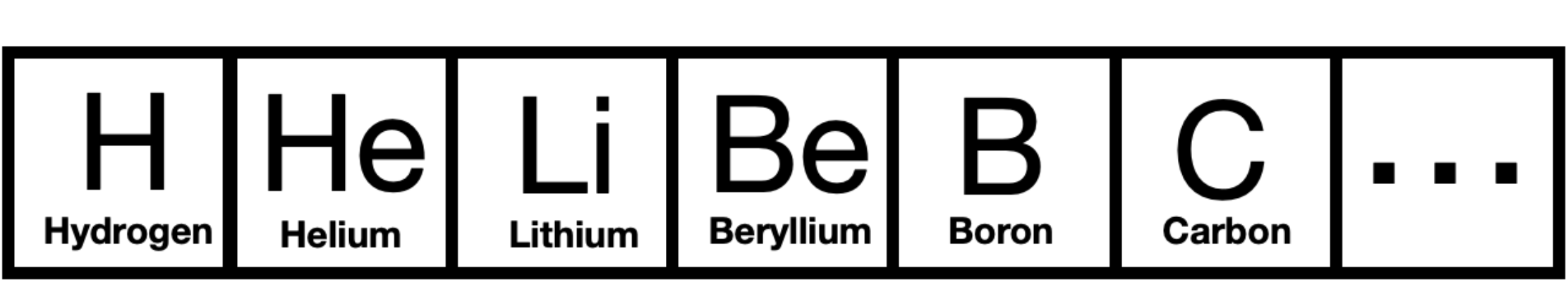}}
        \end{minipage} \\
    \hline
  \end{tabular}
  \caption{Comparison of Structure-based GNNs and Contrastive Knowledge-aware GNN. GNNs without domain knowledge can be fragile when applying to chemical domain since small changes in structure leads to chemicals with different properties.}
\label{reaction}
\end{table*}

To further capture domain knowledge from chemical structures implicitly in a self-supervised manner, we propose to couple GNNs with contrastive learning framework. Contrastive learning compares the differences among a batch of molecules each time and builds domain knowledge from trained molecules (e.g. the reactants and products of the halogenation reaction in Table \ref{reaction}), thus the model is aware of atom or functional group changes. Besides, compared to natural language processing, using contrastive learning helps GNNs to learn molecular representations with less ambiguity from much smaller atom vocabulary and highly frequent atoms (words), resulting in a model with stronger generalization for downstream tasks.

In this study, we propose the Contrastive Knowledge-aware GNN (CKGNN) which learns from large amounts of (over 2.4 million) molecules under a self-supervised contrastive learning framework to produce their corresponding molecular embeddings in a latent space, preserving the fingerprint similarity and domain knowledge of molecules. These molecular embeddings can be later used in downstream tasks such as molecular property predictions. We evaluated the proposed method on eight benchmark datasets, observing a 6\% absolute improvement on average. Further ablation study and analysis also showed that the key to our improvement comes from leveraging domain knowledge explicitly. The main contributions are as follows.
\begin{itemize}
    \item We propose a knowledge-aware GNN that explicitly encodes subgraph-level domain knowledge into molecular embeddings, outperforming previous pre-trained models.
    \item We introduce molecule cluster strategy under a contrastive learning framework, leveraging fingerprints and functional groups as chemical domain knowledge to train knowledge-aware GNN in a self-supervised manner.
    \item We conduct extensive experiments on eight public benchmark datasets of molecular property prediction, demonstrating the effectiveness of our CKGNN model.  
\end{itemize}

\section{Related Work}\label{related work}
\paragraph{Molecular representation learning}
The goal of molecular representation learning is to embed each atom in a molecule into a low-dimensional vector space. Traditional methods utilize descriptors (e.g. ECFPs) to encode the neighbors of atoms in a compound into a fixed bit string with a hash function. Molecular linear notation (e.g. SMILES) is another molecular representation method that abstracts molecular topological information based on common chemical bonding rules. \cite{duvenaud2015convolutional} first applied convolutional layers to map molecules into neural fingerprints. Other attempts feed SMILES into more complicated neural networks to produce molecular representations\cite{zheng2020predicting}. To better capture expressive information, several works also focus on message passing framework to model atom and bond interactions\cite{klicpera2020directional}. Most of them suffered from information redundancy during iterations\cite{song2020communicative}.

\paragraph{Contrastive learning}
The main idea of contrastive learning is to learn discriminative representations by contrasting positive and negative examples\cite{li2019graph}. Contrastive learning has started gaining popularity in several fields in recent years. In natural language processing, INFOXLM\cite{chi2020infoxlm} proposes a cross-lingual pre-training model, which differentiates machine translation of input sequences by contrastive learning. In computer vision, several works\cite{he2020momentum} learn self-supervised image representations by evaluating contrastive loss between differently augmented views. For graph data, traditional methods try to generate negative samples by randomly shuffling node features\cite{velickovic2019deep,hassani2020contrastive}, removing edges or masking nodes\cite{zhu2020deep}. However, these perturbations can hurt the semantic information and domain knowledge of graphs. Especially for chemical compounds, a removal or addition of a covalent bond can drastically change their identities and properties.

\paragraph{Pre-training on molecular graphs}
Despite the fruitful progress of pre-training for convolutional neural networks (CNNs), the counterparts for molecular graphs remain hard challenges. Several works attempt different ways of pre-training GNNs with supervised tasks on molecular datasets. However, models trained on labeled data only involve knowledge in these specific tasks, and the lack of labeled data in molecular tasks also limits the application of models in actual scenarios. \cite{hu2019strategies,hu2020gpt} also propose context prediction and node/edge attributes masking tasks for self-supervised pre-training. Since there are few types of atoms and bonds in molecules, and some atoms (e.g. Carbon and Hydrogen) appear in almost every molecule, the trained model may not be able to capture the valuable chemical domain information\cite{rong2020self}. Inspired by previous works, GROVER\cite{rong2020self} improves the model by introducing semantic motif prediction task. Nevertheless, all these works only implicitly incorporate domain knowledge.

\begin{figure*}[!ht] 
\centering 
\includegraphics[width=1.0\textwidth]{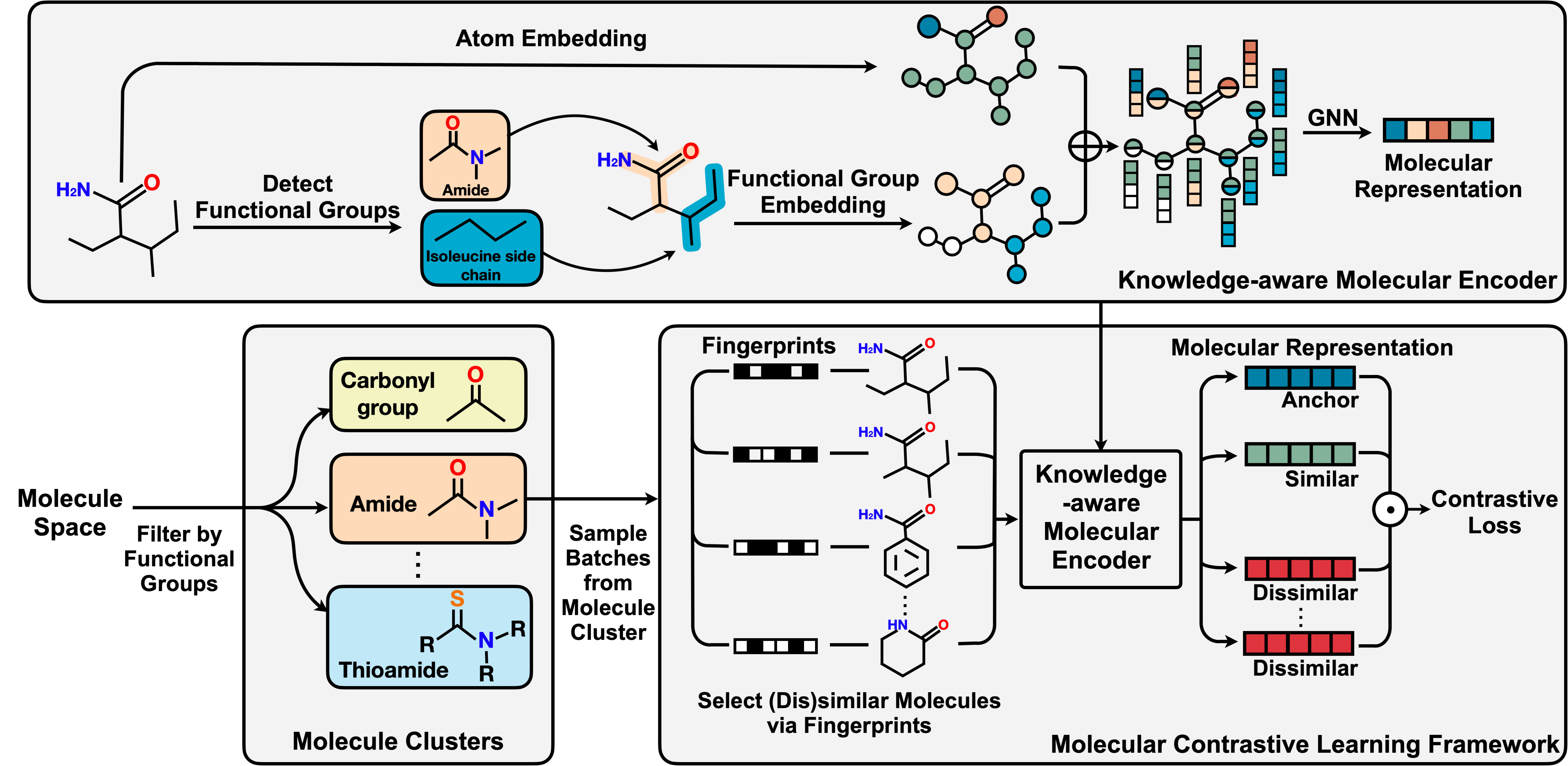}
\caption{Overview of CKGNN Architecture. } 
\label{framework}
\vspace{-0.4cm}
\end{figure*}

\section{Preliminaries}\label{preliminaries}
\subsection{Molecular Graph}

A molecular graph is a heterograph for representing the structural formula of a molecule, whose nodes and edges correspond to the atoms and chemical bonds of the molecule respectively. The atoms are labeled with the types of corresponding atoms and edges are labeled with types of bonds. A hydrogen-suppressed molecular graph is the molecular graph with hydrogen nodes deleted. Since a chemical bond is an attraction between atoms, a molecule graph can be further simplified as a pure graph with types of edges omitted. We observe that hiding the types of edge information does not harm model performance. We use the simplified hydrogen-suppressed molecular graphs for the rest of our paper, which will be defined formally in section \ref{model}.

\subsection{Chemical Domain Knowledge}
Chemical domain knowledge comes from the accumulation of long-term experience of experts. We leverage two types of domain knowledge into our model, including fingerprints and functional groups. Fingerprints are a type of global domain knowledge encoding the structure of molecules. Functional groups are a type of local domain knowledge clustering sets of atoms (nodes) as connected components.

\paragraph{Fingerprints}
Fingerprints encode the neighbors of atoms in a molecule as a fixed length binary vector. Different kinds of fingerprints are developed according to different strategies. There are two major types including structural keys (MACCS) and hashed fingerprints (ECFPs and RDKit Fingerprints). Structural keys encode the structure of molecules as a binary bit string, where each bit corresponds to a predefined structural feature. Hashed fingerprints are generated by enumerating all possible fragments not bigger than a specific size in the molecule, and then converting them into numeric values with a hash function.

\paragraph{Functional Groups}
A functional group is a group of atoms in a molecule with distinctive chemical properties regardless of the rest of the molecule's composition. This type of chemical domain knowledge enables systematic prediction of chemical reactions and the design of chemical synthesis (e.g. drug synthesis). Functional groups can be treated as a subclass of frequent subgraphs which the number of occurrences exceeds a certain thresholds. It is deemed interesting to further investgate frequent groups of atoms in molecule datasets \cite{mrzic2018grasping}, generalizing the concept of functional groups to frequent subgraphs, where domain knowledge can be extracted automatically.

\section{The CKGNN}\label{model}
In this section, we present Contrastive Knowledge-aware Graph Neural Network (CKGNN) on molecular graphs. Section \ref{encoder} explains the core knowledge-aware molecular encoder in details. Section \ref{4.2} presents an overview of our model including molecule cluster strategy for data selection and molecular contrastive learning framework for training our model.

\subsection{Knowledge-aware Molecular Encoder}\label{encoder}
At the core of CKGNN, we develop a knowledge-aware molecular encoder $g$ to generate fixed-size embedding for each molecule by the aggregation of domain knowledge at both atom-level and subgraph-level, as shown in the upper part of Figure \ref{framework}.

For each molecule, we detect functional groups following the largest-first rule, meaning the large functional groups are matched before the smaller functional groups. We observe 99\% coverage of functional groups for all datasets using a pre-defined collection of functional groups from DayLight \footnote{\url{https://www.daylight.com/dayhtml_tutorials/languages/smarts/smarts_examples.html\#GROUP}}. For each atom $v$ in a molecule $G=\{\mathcal{V}, \mathcal{E}\}$, the initial node feature $\boldsymbol{v}$ is two-fold. One is atom embedding $\boldsymbol{v}_{atom}$ which assigns each type of atoms with a one-hot embedding. The other is one-hot functional group embedding $\boldsymbol{v}_{fg}$ indicating which functional group it belongs to. Two embedding layers $\boldsymbol{E}_{atom}$ and $\boldsymbol{E}_{fg}$ are initialized with standard normal distribution. We compute node features as

\begin{equation}
    \boldsymbol{v^{(0)}} = \operatorname{CONCAT}[\boldsymbol{v}_{atom}\boldsymbol{E}_{atom}; \boldsymbol{v}_{fg}\boldsymbol{E}_{fg}]
\end{equation}

Injecting function group embeddings to each atom distinguishes the same types of atoms in different functional groups, easing the issue of small vocabulary and highly frequent atoms. Note that atoms with no functional group matches are marked as \textbf{NOTFOUND} with a zero-padding embedding.

We then adopt common GNN encoders including GCN, GraphSAGE and GIN to aggregate and transform the neighborhood of the input node features of molecule graphs. Formally, the node representation $\boldsymbol{v^{(l)}}$ at the $l$-th layer can be formulated as:

\begin{equation}
    \boldsymbol{v}^{(l)} =\operatorname{COMBINE}^{(l)}(\boldsymbol{v}^{(l-1)}, \operatorname{AGG}^{(l)}(\{\boldsymbol{u}^{(l-1)}|\boldsymbol{u} \in \mathcal{N}(v)\}))
\end{equation}

\noindent where $\mathcal{N}(v)$ is the set of neighbors of node $v$.

A READOUT function is applied to pool node features at the final layer $L$

\begin{equation}
    \boldsymbol{h} = \operatorname{READOUT}(\{\boldsymbol{v}^{(L)}|v \in \mathcal{V}\})
\end{equation}

\noindent where READOUT is the average pooling function. 

The intuition behind the knowledge-aware molecular encoder is that explicitly fusing domain knowledge such as functional groups into GNNs introduces strong inductive bias to our knowledge-aware molecular encoder. We observe that, in experiment section, molecular representation learning at atom-level only is not enough for mastering chemical domain knowledge. To empower our model, we explicitly encode subgraph-level domain knowledge into the process of molecular representation learning.

\subsection{Molecular Contrastive Learning Framework}\label{4.2}

\paragraph{Molecule Cluster Strategy} To simplify the data selection process of contrastive learning, we use molecule cluster strategy to filter all molecules into different molecule clusters. We assign each molecule with a cluster id based on the largest functional group matched. We cluster all molecules into 43 clusters. Clusters with a frequency less than 2,000 are truncated and regrouped into a single cluster. 

Within each molecule cluster, we randomly sample a minibatch of $N$ molecules. We do not sample negative examples explicitly. As shown in the lower part of Figure \ref{framework}, we treat each molecule as the anchor and select the most similar molecule in a batch as the similar instance and leave the rest molecules as $N-2$ dissimilar instances. The goal of molecular contrastive learning framework is to learn from the molecular fingerprint space and build chemical domain knowledge understanding with the output embedding space. For fingerprint similarity computation, we use Dice similarity coefficient.

\paragraph{Loss Function}

We use contrastive learning as a self-supervised pre-training task of knowledge-aware molecular encoder.

In a minibatch of size $N$, the representations of the anchor molecule, the similar instance and dissimilar instances are denoted $\boldsymbol{h}_0$, $\boldsymbol{h}_1^{+}$ and $\{\boldsymbol{h}_{2}^{-},\boldsymbol{h}_{3}^{-},\dots,\boldsymbol{h}_{N-1}^{-}\}$. Note that we encode each molecule instance $G_i$ into fixed-size dense representations $h_i$ with the above mentioned knowledge-aware graph encoder.

\begin{equation}
    \boldsymbol{h}_i = g(G_i)
\end{equation}

The final loss function is computed as:
\begin{equation}
    \mathcal{L}=-\log \frac{\exp \left(\boldsymbol{h}_0^{\top} \boldsymbol{h}_1^{+} / \tau\right)}{\exp \left(\boldsymbol{h}_0^{\top} \boldsymbol{h}_1^{+} / \tau\right) +\sum_{i=2}^{N-1} \exp \left(\boldsymbol{h}_0^{\top} \boldsymbol{h}_{i}^{-} / \tau\right)}
\end{equation}

\noindent where $\tau$ denotes a temperature parameter.


\begin{table*}[!ht]
\renewcommand\arraystretch{1.3}
\centering
\begin{tabular}{c|p{0.3cm}<{\centering}|p{0.3cm}<{\centering}|p{0.9cm}<{\centering}p{0.9cm}<{\centering}p{0.9cm}<{\centering}p{0.9cm}<{\centering}p{0.9cm}<{\centering}p{0.9cm}<{\centering}p{0.9cm}<{\centering}p{0.9cm}<{\centering}p{0.9cm}<{\centering}}
\hline
\multicolumn{3}{c|}{Dataset}  & BBBP & Tox21 & SIDER & ClinTox & MUV & HIV & BACE & Toxcast & Average\\
\hline
\multicolumn{3}{c|}{\# Molecules}     & 2039  & 7831 & 1427 & 1478 & 93087 & 41127 & 1513 & 8575  &/  \\
\multicolumn{3}{c|}{\# Binary prediction tasks}   & 1  & 12 & 27 & 2 & 17 & 1 & 1 & 617 &/   \\
\hline
Method & PT & CL &  & &  &  & &  &  & & \\
\hline
GCN & \XSolidBrush & \XSolidBrush & 0.6489 & 0.7368 & 0.6010 & 0.6281 & 0.7298 & 0.7576 & 0.7261 & 0.6291 &0.6822\\
GIN & \XSolidBrush & \XSolidBrush   &0.6286 &0.7421 &0.5647 &0.5533 &0.7160 &0.7498 &0.7476 &0.6302 &0.6665\\
GraphSAGE & \XSolidBrush & \XSolidBrush &0.6956 &0.7472 &0.6032 &0.5687 &0.7256 &0.7486 &0.7243 &0.6356 &0.6811\\
\hline
Pre-trained GCN & \Checkmark & \XSolidBrush &0.7051 &0.7552  &0.6256 &0.6323 &0.7938 &0.7845 &0.8244 &0.6523 &0.7217\\
Pre-trained GIN & \Checkmark & \XSolidBrush &0.6867 &0.7857 &0.6274 &0.7258 &0.8139 &0.7990 &0.8457 &0.6576 &0.7427\\
Pre-trained GraphSAGE & \Checkmark & \XSolidBrush &0.6391 &0.7683 &0.6075 &0.6072 &0.7842 &0.7621 &0.8079 &0.6489 &0.7032\\
GCC \cite{qiu2020gcc} & \Checkmark & \Checkmark &0.7797  &0.7112  &0.5979  &0.6813 &0.6065 &0.6582  &0.6629 &0.6906 &0.6735\\
\hline
KNN & \XSolidBrush & \XSolidBrush    & 0.8635 & 0.7529   & 0.6432 &0.7622 &0.6320 &0.8261 &0.8575 &0.6403 &0.7472\\
\hline
CKGNN(GCN)   & \Checkmark & \Checkmark  & 0.8886  & 0.8179 & 0.6632 & \textbf{0.8286} & 0.8267 & 0.8214 & 0.8854 & 0.7238   &0.8070 \\
CKGNN(GIN)  & \Checkmark & \Checkmark & 0.8986  & \textbf{0.8263} & \textbf{0.6779} & 0.7781 & 0.8297 & 0.8284 & 0.8881 & \textbf{0.7351}    &0.8078\\
CKGNN(GraphSAGE) & \Checkmark & \Checkmark & \textbf{0.9072}  & 0.8126 & 0.6598 & 0.7813 & \textbf{0.8333} & \textbf{0.8414} & \textbf{0.9043} & 0.7279  & \textbf{0.8085} \\
\hline
\end{tabular}
\caption{Test ROC-AUC Performance Comparison on Molecular Prediction Benchmark. PT is short for pre-training. CL is short for contrastive learning.}
\label{result}
\vspace{-0.4cm}
\end{table*}

\section{Experiments}\label{experiments}

In this section, we conduct extensive experiments on molecular property prediction tasks to evaluate the superiority of our proposed CKGNN. We begin by introducing our experimental setups including datasets, baselines, and evaluation protocols. We then compare CKGNN with prior state-of-the-art baselines to demonstrate the effectiveness of CKGNN. We also conduct ablation study to investigate the validity of key components. We further discuss the quality of molecular embeddings generated by our CKGNN, revealing the importance of chemical domain knowledge in pre-trained models.

\subsection{Experimental Setups}\label{setup}
\paragraph{Datasets}
We collect 2.4 million unlabeled molecules sampled from ZINC15 \cite{sterling2015zinc} and Chembl \cite{gaulton2012chembl} for self-supervised pre-training. To evaluate the effectiveness of CKGNN, we conduct experiments on eight datasets from the public benchmark MoleculeNet \cite{wu2018moleculenet}, including BBBP, Tox21, SIDER, ClinTox, MUV, HIV, BACE and Toxcast for classification tasks. The statistics of datasets are shown in Table \ref{result}.

\paragraph{Baselines}
As shown in Table \ref{result}, we compare our model with eight baselines in three groups. In the first group, GCN \cite{kipf2016semi}, GIN\cite{xu2018powerful}, GraphSAGE\cite{hamilton2017inductive} predict molecular property without pre-training. The second group is the pre-training group, including pre-training GCN, GIN, GraphSAGE and Graph Contrastive Coding (GCC, \cite{qiu2020gcc}). The third group is k-nearest neighbor algorithm (kNN) with $k=10$ based on molecular fingerprint similarity.

\paragraph{Implementation details}
We use 512-dimensional embeddings for both atom embeddings and function group embeddings. We use two-layer GNN in all experiments with hidden dimension of 512 followed by 256. We use the Adam optimizer to train all models with a learning rate of 0.00001 and batch size of 32 for 5 epoches. We implement our model using PyTorch and Deep Graph Library. The temperature parameter is set to 0.07.

\paragraph{Evaluation protocol}
To evaluate knowledge-aware molecular representations, we apply multiple layer perceptron (MLP) on top of generated molecular embeddings to predict the property of the molecular graph. Note that we freeze the parameters and fix the generated embeddings. We evaluate test performance on downstream tasks using ROC-AUC.

\subsection{Results}

Table \ref{result} shows the overall results of eight baselines on eight classification datasets. The first six rows are taken from \cite{hu2019strategies}. The bold numbers indicate the best results achieved by CKGNN with three different GNNs as graph encoder.

Comparing the first two groups in the first seven rows, it seems that pre-training with large amounts of data and massive GPU hours leads to substantive improvement, as known in computer vision and natural language processing. We show that kNN $(k=10)$ with molecular fingerprint similarity outperforms pre-training models based on graph structure information only while suppressed by our CKGNN, indicating chemical domain knowledge is a supplement to pre-training models.

CKGNN also outperforms GCC which adopts contrastive learning tasks in pre-training framework, revealing the fusion of chemical domain knowledge and pre-training technique is the best of both worlds.

\subsection{Ablation Studies}

To investigate the contribution of different factors that influence the performance of CKGNN, we conducted ablation studies on the five downstream datasets. As shown in Table \ref{ablation}, decoupling knowledge-aware features in molecular encoder from contrastive learning framework yields a significant drop of performance. Molecule cluster strategy is also crucial and effective for better results. Different types of fingerprints influence the overall performance slightly. This evidence supports our claims that injecting chemical domain knowledge into contrastive learning framework improves model performance.

\begin{table}[!ht]
\vspace{-0.2cm}
\renewcommand\arraystretch{1.1}
\centering
\begin{tabular}{p{2cm}|p{0.8cm}<{\centering}p{0.8cm}<{\centering}p{0.8cm}<{\centering}p{0.8cm}<{\centering}p{0.8cm}<{\centering}}
\hline
                & BBBP     &  HIV    & BACE    & ClinTox & MUV     \\
\hline
Ours(RDKit)     & 0.9072   & 0.8414  & 0.9043  & 0.7813  & 0.8333  \\
\hline
--KA            & 0.8881   & 0.7954  & 0.8728  & 0.7408  & 0.7989  \\
--MC            & 0.8797   & 0.8331  & 0.8906  & 0.7755  & 0.8215  \\
--KA \& MC      & 0.8542   & 0.7914  & 0.8028  & 0.7267  & 0.7975  \\
\hline
Ours(MACCS)     & 0.8983   & 0.8235  & 0.8890  & 0.8168  & 0.8078  \\
Ours(morgan)    & 0.9001   & 0.8252  & 0.8913  & 0.8048  & 0.7942  \\
\hline
\end{tabular}
\caption{Ablation studies. KA is short for knowledge-aware features in our molecular encoder. MC is short for molecule cluster strategy. RDKit, MACCS and morgan correspond to different fingerprints.}
\label{ablation}
\end{table}

\subsection{Similarity Distribution Analysis}

To compare the similarity distributions produced by different models, we randomly sample 25 molecules as anchors and 100 molecules for each anchor, resulting in 2500 sample pairs. We compute the similarity score of all pairs using molecular fingerprint, GCC, pre-trained GIN and CKGNN. 

Note that GCC achieves good results in fine-tuning stage but its similarity distribution is squeezed in the range of 0.75 to 0.99.

From a variance perspective, the similarity distribution of CKGNN has the largest variance of 0.067 among four distributions, indicating that injecting chemical domain knowledge improves the diversity of molecular representations and much more information is stored in the output space of molecule embeddings.

\begin{figure}[!ht]
\centering 
\includegraphics[width=0.45\textwidth]{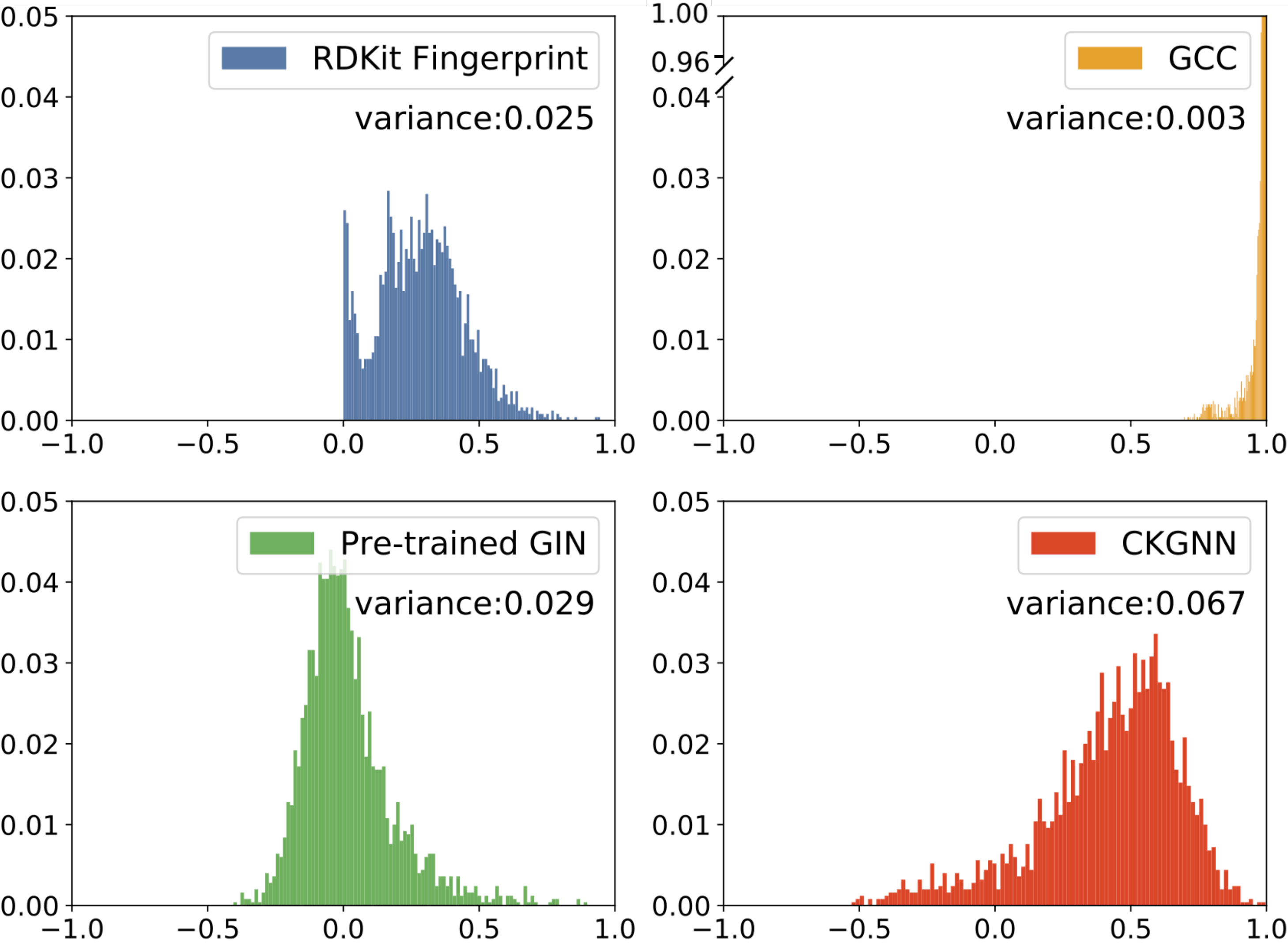}
\caption{Similarity distribution of molecular fingerprint, GCC, pre-trained GIN and CKGNN.} 
\label{correlation} 
\end{figure}

\subsection{Case Study: The Most Similar Molecules}

After randomly selecting a molecule as anchor, we find the most similar molecules in the whole molecule space by ranking embedding similarity score. We observe that only CKGNN can successfully identify molecules with the same substructures, including a special functional group in orange (Carboxyl group) and a frequent substructure in yellow (Benzene ring). GNN purely based on structure information is hard to master chemical domain knowledge, leaving the burden of explaining prediction.

\begin{table}[!ht]
\renewcommand\arraystretch{1.5}
  \centering
  \begin{tabular}{p{1.2cm}<{\centering}p{1.2cm}<{\centering}p{1.8cm}<{\centering}p{2cm}<{\centering}}
\toprule
    Method &Anchor& Top-1& Top-2 \\
   \midrule
    CKGNN&\begin{minipage}[b]{0.17\columnwidth}
  \centering
  \raisebox{-.3\height}{\includegraphics[width=\linewidth]{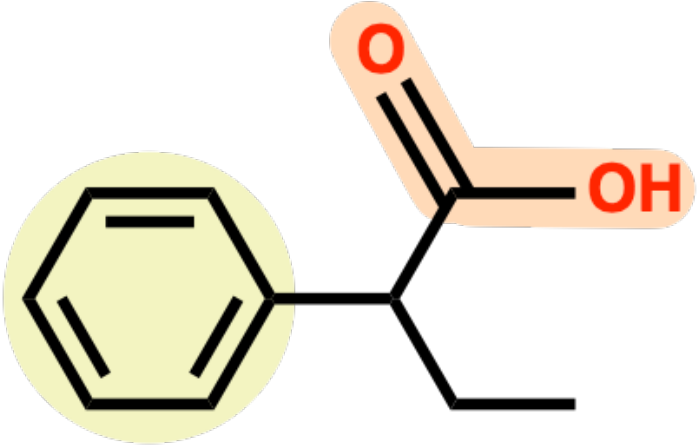}}
 \end{minipage}
    & 
    \begin{minipage}[b]{0.25\columnwidth}
  \centering
  \raisebox{-.3\height}{\includegraphics[width=\linewidth]{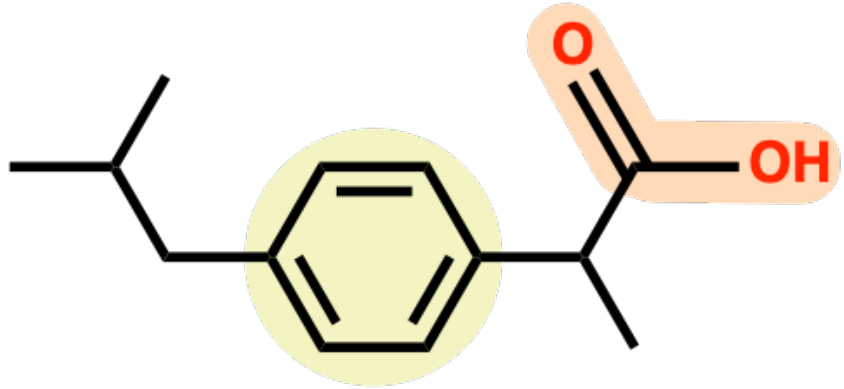}}
 \end{minipage}
    & \begin{minipage}[b]{0.26\columnwidth}
  \centering
  \raisebox{-.3\height}{\includegraphics[width=\linewidth]{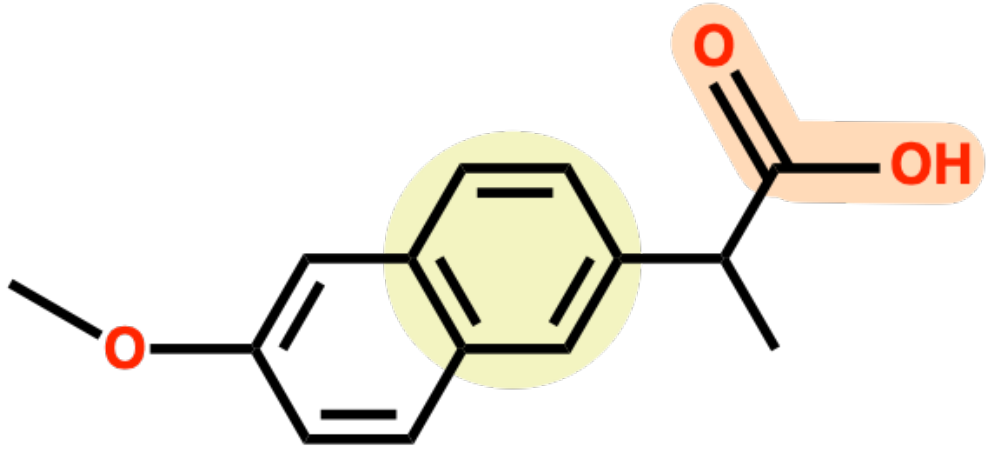}}
 \end{minipage} \\
    GCC&\begin{minipage}[b]{0.17\columnwidth}
  \centering
  \raisebox{-.3\height}{\includegraphics[width=\linewidth]{fig/anchor.pdf}}
 \end{minipage}
    & 
    \begin{minipage}[b]{0.17\columnwidth}
  \centering
  \raisebox{-.3\height}{\includegraphics[width=\linewidth]{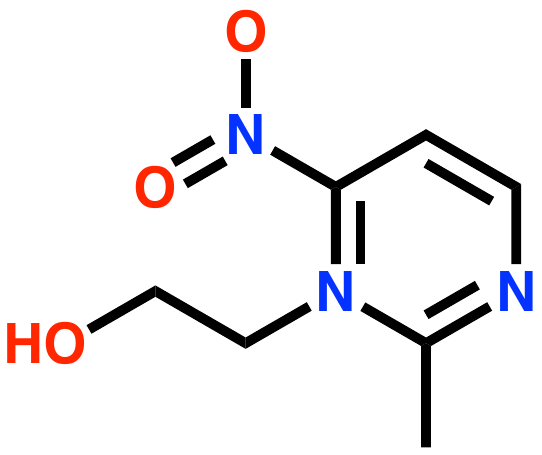}}
 \end{minipage}
    & \begin{minipage}[b]{0.24\columnwidth}
  \centering
  \raisebox{-.3\height}{\includegraphics[width=\linewidth]{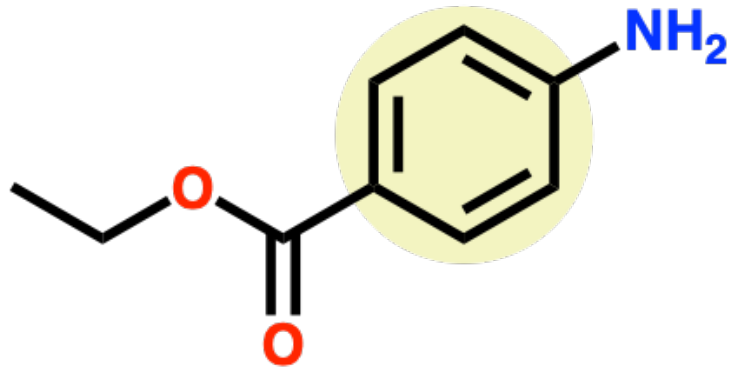}}
 \end{minipage} \\
    Pre-trained GIN&\begin{minipage}[b]{0.17\columnwidth}
  \centering
  \raisebox{-.3\height}{\includegraphics[width=\linewidth]{fig/anchor.pdf}}
 \end{minipage}
    & 
    \begin{minipage}[b]{0.2\columnwidth}
  \centering
  \raisebox{-.8\height}{\includegraphics[width=\linewidth]{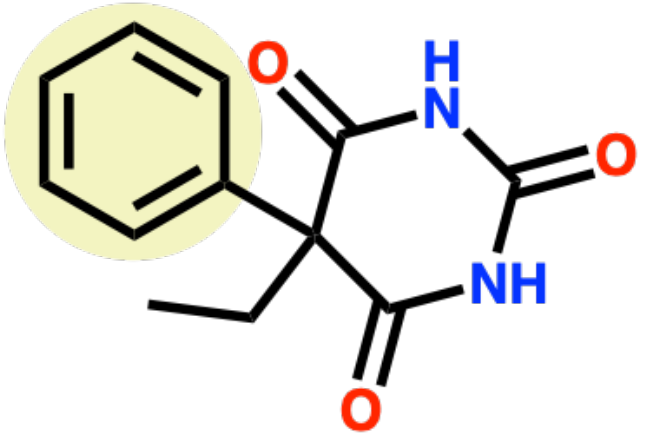}}
 \end{minipage}
    & \begin{minipage}[b]{0.2\columnwidth}
  \centering
  \raisebox{-.7\height}{\includegraphics[width=\linewidth]{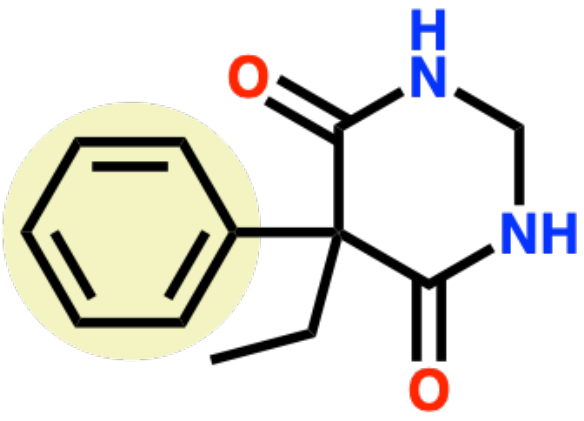}}
 \end{minipage} \\
\bottomrule
  \end{tabular}
  \caption{The most similar molecules seleted by CKGNN, GCC and Pretrained GIN.}
\label{similar}
\end{table}

\subsection{Case Study: Functional Isomerism}
Functional isomers are structural isomers which have different functional groups, resulting in significantly different chemical and physical properties. Table \ref{isomer} shows that GCC and Pre-trained GIN produce almost inseparable features for molecules with different functions and properties while CKGNN sets a much lower similarity score for the same molecule pairs. These results support the superiority of the fusion of chemical domain knowledge and pre-training technique.

\begin{table}[!ht]
\renewcommand\arraystretch{1.5}
  \centering
    \begin{tabular}{p{1cm}<{\centering}p{1cm}<{\centering}ccc}
    \toprule
    \multicolumn{2}{c}{Molecule pairs}& CKGNN& GCC &P-GIN \\
    \midrule
    \begin{minipage}[b]{0.12\columnwidth}
  \centering
  \raisebox{-.2\height}{\includegraphics[width=\linewidth]{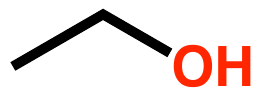}}
 \end{minipage}
    & 
    \begin{minipage}[b]{0.12\columnwidth}
  \centering
  \raisebox{-.2\height}{\includegraphics[width=\linewidth]{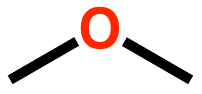}}
 \end{minipage}
    & \ 0.2661 & 0.9999 &0.9121\\
   \begin{minipage}[b]{0.13\columnwidth}
  \centering
  \raisebox{-.2\height}{\includegraphics[width=\linewidth]{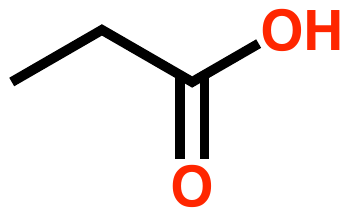}}
 \end{minipage}
    & 
    \begin{minipage}[b]{0.13\columnwidth}
  \centering
  \raisebox{-.2\height}{\includegraphics[width=\linewidth]{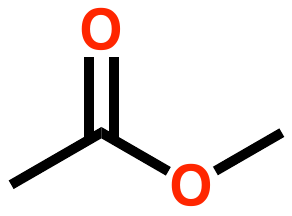}}
 \end{minipage}
    & \ 0.1180 &0.9998 &0.8931\\
    \begin{minipage}[b]{0.13\columnwidth}
  \centering
  \raisebox{-.2\height}{\includegraphics[width=\linewidth]{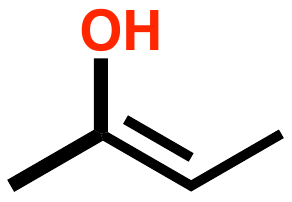}}
 \end{minipage}
    & 
    \begin{minipage}[b]{0.13\columnwidth}
  \centering
  \raisebox{-.2\height}{\includegraphics[width=\linewidth]{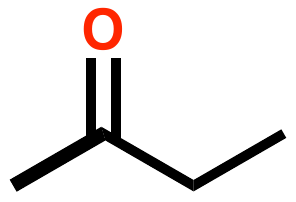}}
 \end{minipage}
    & \ 0.2512 &0.9999 &0.8527 \\
    \begin{minipage}[b]{0.13\columnwidth}
  \centering
  \raisebox{-.2\height}{\includegraphics[width=\linewidth]{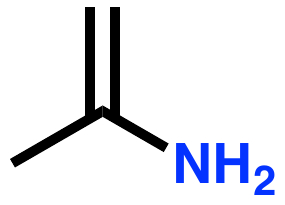}}
 \end{minipage}
    & 
    \begin{minipage}[b]{0.13\columnwidth}
  \centering
  \raisebox{-.2\height}{\includegraphics[width=\linewidth]{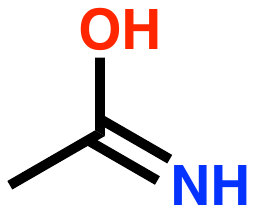}}
 \end{minipage}
    & -0.3693 &0.9999 &0.9821 \\
 \begin{minipage}[b]{0.14\columnwidth}
  \centering
  \raisebox{-.3\height}{\includegraphics[width=\linewidth]{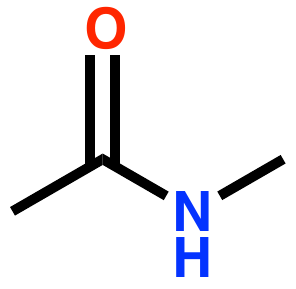}}
 \end{minipage}
    & 
    \begin{minipage}[b]{0.14\columnwidth}
  \centering
  \raisebox{-.3\height}{\includegraphics[width=\linewidth]{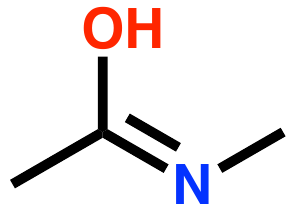}}
 \end{minipage}
    & -0.4476 &0.9912 &0.8793 \\
  \bottomrule
  \end{tabular}
  \caption{Embedding similarity on five pairs of functional isomers using three different models. P-GIN is short for pre-trained GIN.}
\label{isomer}
\end{table}

\section{Conclusion}\label{conclusion}
We propose the Contrastive Knowledge-aware GNN (CKGNN) for self-supervised molecular representation learning. Crucial to the success of our model is to consider both chemical domain knowledge and molecular structure information in combination with a contrastive learning framework. This ensures that the generated molecular embeddings capture chemical domain knowledge to distinguish molecules with similar chemical formula but dissimilar functions, which are useful for achieving good results on downstream tasks. Experiments on multiple datasets, diverse downstream tasks and different GNN architectures support our claim that the fusion of chemical domain knowledge achieves consistently better than models based on domain knowledge or large amount of data solely.

It is also interesting to further investigate automatic extraction of chemical domain knowledge such as frequent subgraphs and incorporate this type of information into molecular representation learning, making our model more knowledgable.

\newpage

\appendix
\bibliographystyle{named}
\bibliography{mcl}

\end{document}